\RequirePackage[hyphens]{url}
\documentclass[amssymb, notitlepage, pra, 12pt,aps,table,nofootinbib]{revtex4-2}
 
\usepackage[utf8]{inputenc}
\usepackage{amsthm}
\usepackage{mathptmx, amsmath}
\usepackage{amsfonts}
\usepackage{mathtools}
\usepackage{physics}
\usepackage{xcolor}
\usepackage{graphicx}
\usepackage{comment}
\usepackage[left=23mm,right=13mm,top=35mm,columnsep=15pt]{geometry} 
\usepackage[colorinlistoftodos, color=green!40, prependcaption]{todonotes}
\usepackage{placeins}
\usepackage[T1]{fontenc}
\usepackage{lipsum}
\usepackage{csquotes}
\usepackage{fancyvrb}
\usepackage{hyperref}
\PassOptionsToPackage{hyphens}{url}
\hypersetup{breaklinks=true}

\usepackage{setspace}
\begin{document}

\title{Dr-COVID: Graph Neural Networks for SARS-CoV-2 Drug Repurposing}

\author{Siddhant Doshi and Sundeep Prabhakar Chepuri}
\thanks{The authors are with the Department of Electrical Communication Engineering, Indian Institute of Science, Bangalore, India. Email:\{siddhantd,spchepuri\}@iisc.ac.in.
}

\begin{abstract}
The \emph{2019 novel coronavirus (SARS-CoV-2)} pandemic has resulted in more than a million deaths, high morbidities, and economic distress worldwide. There is an urgent need to identify medications that would treat and prevent novel diseases like the 2019 coronavirus disease (COVID-19). Drug repurposing is a promising strategy to discover new medical indications of the existing approved drugs due to several advantages in terms of the costs, safety factors, and quick results compared to new drug design and discovery. In this work, we explore computational data-driven methods for drug repurposing and propose a dedicated graph neural network (GNN) based drug repurposing model, called \texttt{Dr-COVID}. Although we analyze the predicted drugs in detail for COVID-19, the model is generic and can be used for any novel diseases. We construct a four-layered heterogeneous graph to model the complex interactions between drugs, diseases, genes, and anatomies. We pose drug repurposing as a link prediction problem. Specifically, we design an encoder based on the scalable inceptive graph neural network (\texttt{SIGN}) to generate embeddings for all the nodes in the four-layered graph and propose a quadratic norm scorer as a decoder to predict treatment for a disease. We provide a detailed analysis of the 150 potential drugs (such as \emph{Dexamethasone}, \emph{Ivermectin}) predicted by \texttt{Dr-COVID} for COVID-19 from different pharmacological classes (e.g., corticosteroids, antivirals, antiparasitic). Out of these 150 drugs, 46 drugs are currently in clinical trials. \texttt{Dr-COVID} is evaluated in terms of its prediction performance and its ability to rank the known treatment drugs for diseases as high as possible. For a majority of the diseases, \texttt{Dr-COVID} ranks the actual treatment drug in the top 15. 
\end{abstract}


\maketitle

\section{Introduction} \label{sec:Intro}
The dreadful pandemic outbreak of the coronavirus disease 2019 (COVID-19) has affected about 56 million people with more than a million deaths worldwide as of November 2020. The June 2020 Global Economic Prospects~\cite{economy} estimated a $5.2$\% downfall in the global gross domestic product (GDP) in 2020 that would lead to the worst economic slowdown in history after the Second World War. The disease affects mammals' respiratory tract and shows symptoms similar to pneumonia, causing mild to severe respiratory tract infections~\cite{CoV_1}. The pathogen that causes COVID-19 belongs to the \emph{Coronaviridae} family, which is a family of enveloped positive-strand RNA viruses that affect mammals, birds, and amphibians. The name coronavirus (CoV) is derived because of the crown-shaped spikes that project from their surface. Coronaviruses are majorly grouped into four genera: \emph{alphacoronavirus}, \emph{betacoronavirus}, \emph{deltacoronavirus}, and \emph{gammacoronavirus}. While \emph{deltacoronaviruses} and \emph{gammacoronaviruses} infect birds, \emph{alphacoronaviruses} and \emph{betacoronaviruses} infect mammals~\cite{CoV_2}. Out of the seven known strains of human CoVs (HCoVs), the three \emph{betacoronaviruses}, namely, \emph{middle east respiratory syndrome coronavirus (MERS-CoV)}, \emph{severe acute respiratory syndrome coronavirus (SARS-CoV)}, and the \emph{novel severe acute respiratory syndrome coronavirus (SARS-CoV-2)} produce severe symptoms. In the past two decades, the world witnessed highly fatal \emph{MERS-CoV} and \emph{SARS-CoV} that led to global epidemics with high mortality. Although the 2003 \emph{SARS-CoV} outbreak was controlled, it infected 8098 individuals and resulted in 774 deaths. As of November 2019, 2494 cases and 855 deaths were reported due to \emph{MERS-CoV}, with the majority in Saudi Arabia~\cite{CoV_2}. In December 2019, similar cases were again reported in Wuhan City, China~\cite{CoV_4}, wherein investigations confirmed it to be the third novel CoV, i.e., \emph{SARS-CoV-2}, which is also referred to as \emph{HCoV-2019}, \emph{2019-nCoV}, or colloquially simply as coronavirus~\cite{CoV_3}. \emph{SARS-CoV-2} being highly contagious, on 30 January 2020, the World Health Organization (WHO) declared it as a public emergency of international concern warning all the countries with vulnerable health care systems~\cite{CoV_5}. 

The current treatment for COVID-19 is completely supportive and symptomatic as there are no specific known medicines. Several research groups around the world are trying to develop a vaccine that would prevent and treat \emph{SARS-CoV-2}. Looking at the current unpredictable trajectory of how the disease spreads and the life cycle of the virus, there is an urgent need to develop preventive strategies against it. Given this strict timeline, a more realistic solution lies in drug repurposing or drug repositioning, which aims to identify new medical indications of approved drugs. Drug repurposing offers several advantages. It has a low risk of failure as the drug has already been approved with less unknown harmful adverse effects. It  reduces the time frame for drug development as the drugs have passed all the pre-clinical trials and safety norms. Finally, compared to the discovery of a new drug, drug repurposing requires less economic investment and puts fewer lives of volunteers (particularly kids) involved in clinical trials at risk~\cite{DR}. Some of the examples of repurposed drugs are \textit{Sildenafil}, which was initially developed as an antihypertensive drug was proved effective in treating erectile dysfunction by Pfizer~\cite{DR}, and \textit{Rituximab} that was originally used against cancer was proved to be effective against rheumatoid arthritis~\cite{DR}, to name a few. Even for COVID-19, drugs like \textit{Remdesivir} (a drug for treating Ebola virus disease), \textit{Chloroquine/Hydroxychloroquine} (antimalarial drugs), \textit{Dexamethasone} (anti-inflammatory drugs) are being repurposed and are under clinical trials as per the International Clinical Trials Registry Platform (ICTRP), which is a common platform maintained by WHO to track the clinical trial studies across the world. 

Drug repurposing involves identifying potential drugs and monitoring their \emph{in vivo}  efficacy and potency against the disease. The most critical step in this pipeline is identifying the right candidate drugs, for which experimental and computational approaches are usually considered. To identify potential drugs experimentally, a variety of chromatographic and spectroscopic techniques are available for target-based drug discovery. Phenotype screening is used as an alternative to target-based drug discovery when the identity of the specific drug target and its role in the disease are not known~\cite{DR}. Recently, computational approaches are receiving attention due to the availability of large biological data. Efficient ways to handle big data has opened up many opportunities in the field of pharmacology. Zitnik, et al.~\cite{CB_data} elaborates several data-driven computational tools to integrate large volumes of heterogeneous data and solve problems in pharmacology such as drug-target interaction prediction (identify interactions between a drug and its target genes), drug repurposing, and drug-drug interaction or side effect prediction, to list a few. Hence this field is known as computational pharmacology. Many standard machine learning (ML) and deep learning (DL) techniques  have been applied in computational pharmacology. Drug-drug interaction was formulated as a binary classification problem and solved using ML techniques like random forest, support vector machines (SVM), and naive bayes~\cite{DDI_ML}, and using DL models like deep multi-layer perceptrons and recurrent neural networks, to name a few. DL techniques often outperform standard ML techniques~\cite{DDI_DL_1,DDI_DL_2}. However, these methods lack the ability to capture the structural information in the data, specifically the connections between different biological entities (e.g., interactions between drugs and genes or between drugs and diseases). A natural and efficient way to represent such structural information is to construct a graph with nodes representing entities like drugs, genes, diseases, etc., and edges representing the complex interactions between these entities. Graph neural networks (GNNs) capture the structural information by accounting for the underlying graph structure while processing the data. Decagon, a GNN-based model designed for predicting the side effects of a pair of drugs has proved its capability by outperforming the non-graph based machine learning models in terms of its prediction performance~\cite{Decagon}. Similarly, drug repurposing has been studied using computational methods such as signature matching methods, molecular docking, and network-based approaches. Recently, network-based and machine learning approaches~\cite{DR_M_1,DR_M_2,DR_M_3,DR_M_4,DR_M_5}, and GNN based approaches~\cite{Network_medicine} and~\cite{Few_shot} have been proposed for drug repurposing.   

In this work, we propose a GNN architecture for COVID-19 drug repurposing called \texttt{Dr-COVID}, which is a dedicated model for drug repurposing. We formulate our problem by constructing a four-layered heterogeneous graph comprising drugs, genes, diseases, and anatomies. We then build a deep learning model to predict the links between the drug and disease entities, where a link between a drug-disease entity suggests that the drug treats the disease. Specifically, \texttt{Dr-COVID} is based on the scalable inceptive graph neural network (\texttt{SIGN}) architecture~\cite{SIGN} for generating the node embeddings of the entities. We propose a quadratic norm scoring function that rank orders the predicted drugs. All the network information and node features are derived from the drug repurposing knowledge graph (DRKG)~\cite{DRKG}. DRKG is a biological knowledge graph compiled using several databases, and comprises entities like drugs, diseases, anatomies, etc., and their connections. We leverage their generic set of low-dimensional embeddings that represent the graph nodes and edges in the Euclidean space for training. We validate \texttt{Dr-COVID}'s performance on the known drug-disease pairs. Although we present the results and analysis for COVID-19, \texttt{Dr-COVID} is generic and is useful for any novel human diseases. From a list of 150 drugs predicted by \texttt{Dr-COVID} for \emph{SARS-CoV-2}, 46 drugs are currently in clinical trials. For a majority of diseases with known treatment, the proposed \texttt{Dr-COVID} model ranks the approved treatment drugs in the top 15, which suggests the efficacy of the proposed drug repurposing model. As we use the \texttt{SIGN} architecture that does many computations beforehand, \texttt{Dr-COVID} is computationally efficient as compared to the other GNN-based methods~\cite{Network_medicine,Few_shot}. Specifically, in contrast to~\cite{Network_medicine} we include additional entities such as anatomies as the side information in our graph. This additional information provides indirect interactions between the disease and gene entities. The norm scorer we design captures correlations between the drug and disease pairs, and as a consequence, the model predicts many more drugs (e.g., \emph{Brexanolone}) that are in clinical trials as compared to the existing GNN-based and network-based drug repurposing models.

\section{Results and discussion} \label{sec:Results}
\begin{figure*}
    \centering
    \includegraphics[height=587pt,width=515pt]{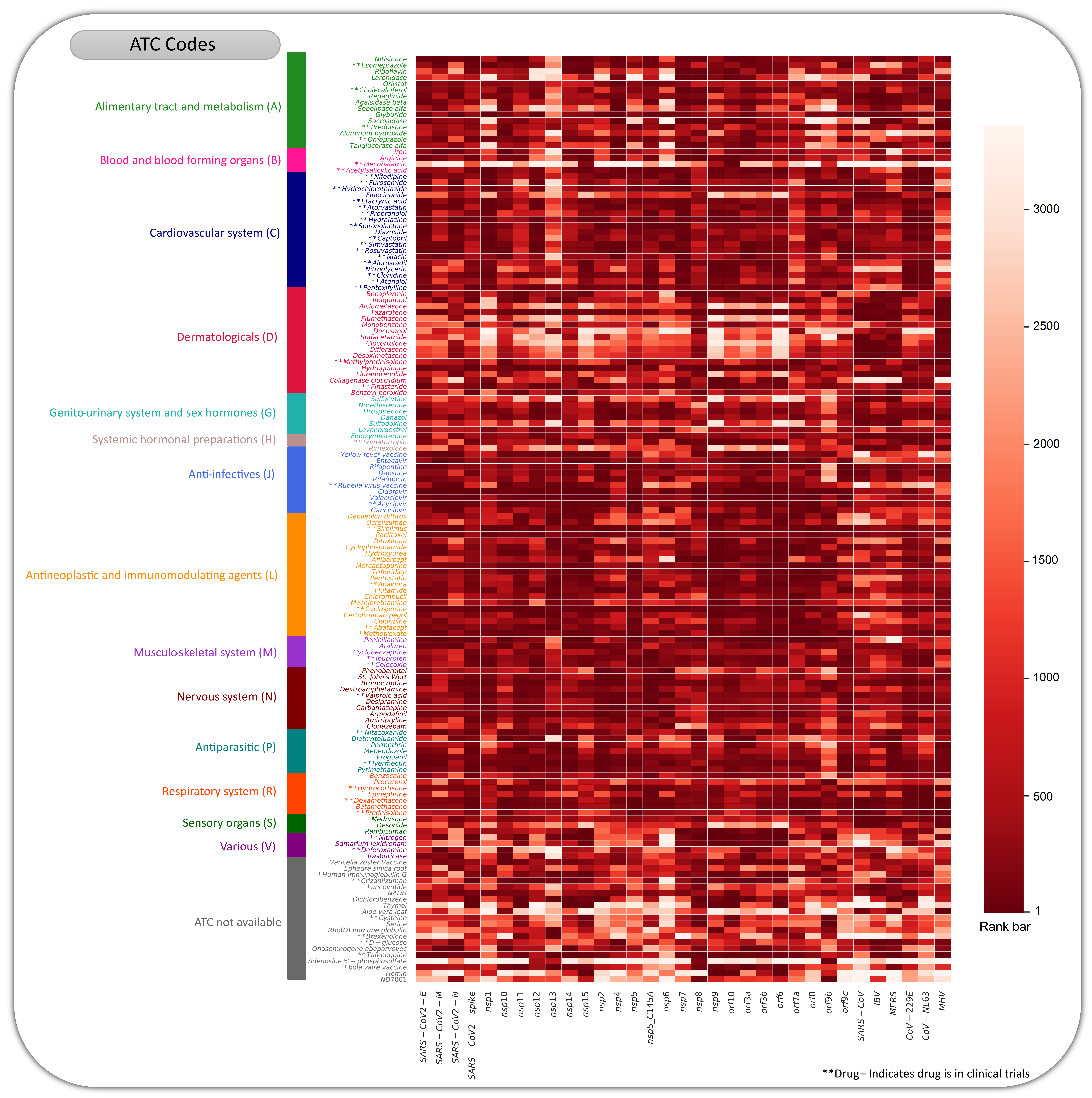}
    \caption{Rank heatmap of the predicted drugs with their corresponding ATC labels.}
    \label{fig:Heatmap}
\end{figure*}

In this section, we present the drugs predicted by \texttt{Dr-COVID} for COVID-19 according to their pharmacological classifications, and elaborate on their roles in treating the disease. We individually predict drugs for the 27 entities that specify the \emph{SARS-CoV-2} genome structure as identified by Gordon et al.~\cite{Gordon}. This genome structure includes structural proteins, namely, envelope (\emph{SARS-CoV2-E}), membrane (\emph{SARS-CoV2-M}), nucleocapsid (\emph{SARS-CoV2-N}), surface (\emph{SARS-CoV2-spike}) proteins, 15 non-structural proteins (nsp), and open reading frames (orf) that encode the accessory proteins. We also predict drugs for 6 diseases related to CoV, namely,  \emph{SARS-CoV}, \emph{Avian infectious bronchitis virus (IBV)}, \emph{MERS-CoV}, \emph{CoV-229E}, \emph{CoV-NL63}, and \emph{Murine coronavirus (MHV)}. We choose the top 10 ranked predicted drugs for all these disease targets, combine them, and present them as a single list of 150 drugs (after removing the duplicate entries). We refer to these 33 (i.e., 27 entities related to the \emph{SARS-CoV-2} genome structure and 6 CoV diseases) as COVID-19 nodes. Out of these 150 drugs, 46 drugs are in clinical trials in different phases. We provide the predicted scores of all the drugs for all the COVID-19 nodes using \texttt{Dr-COVID} in our github repository. The software to reproduce the results are available at: \url{https://github.com/siddhant-doshi/Dr-COVID}

Fig~\ref{fig:Heatmap} gives a heatmap indicating the ranks of these 150 drugs. It is a matrix representation in which the drugs are listed on the vertical axis and COVID-19 nodes on the horizontal axis. All the 150 drugs are grouped based on their first-level anatomical therapeutic chemical (ATC) codes as indicated on the left side. A colored patch in the heatmap indicates the rank of a drug for a disease. The darker the patch, the better is the rank, as indicated by the rank bar on the right side. As can be seen, a major portion of the heatmap is covered with dark patches as we only consider the top 10 ranked drugs. We can infer from the heatmap that cardiovascular drugs (e.g., \emph{Captopril}, \emph{Atenolol}) and anti-inflammatory drugs (e.g., \emph{Celecoxib}, \emph{Prednisone}) are ranked high for the \emph{alphacoronaviruses}, and a combination of antiparasitic (e.g., \emph{Ivermectin}), corticosteroids (e.g., \emph{Prednisolone}, \emph{Dexamethasone}), antivirals (e.g., \emph{Cidofovir}), and antineoplastic drugs (e.g., \emph{Methotrexate}, \emph{Sirolimus}) in the case of \emph{betacoronaviruses}.

\begin{figure*}
    \centering
    \includegraphics[scale=0.4]{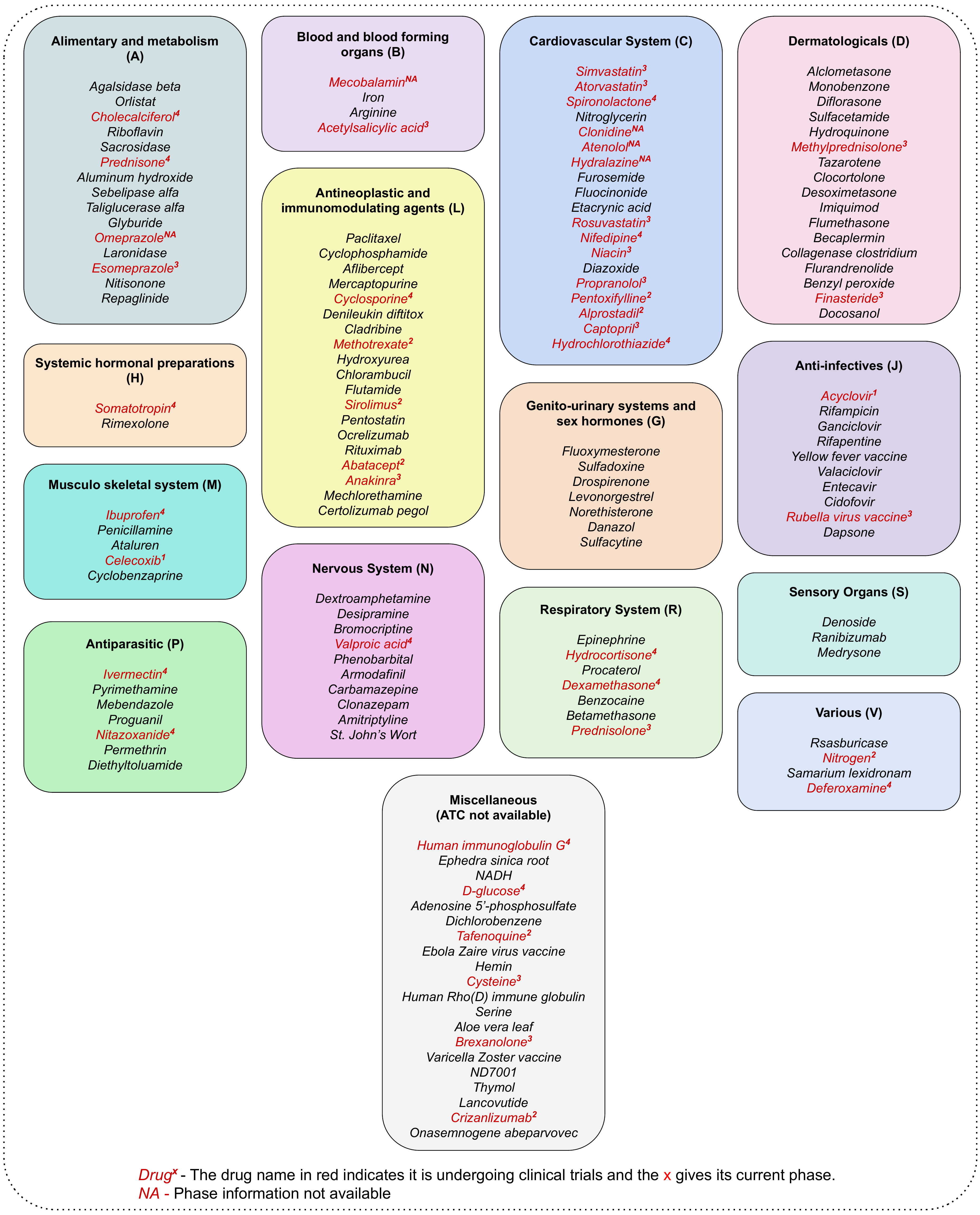}
    \caption{Predicted drugs for COVID, categorized based on their ATC labels.}
    \label{fig:Just_list}
\end{figure*}

Fig~\ref{fig:Just_list} lists the drugs predicted by \texttt{Dr-COVID}, grouped based on their first level ATC codes such as antiparasitic (P), respiratory system (R), and so on, whereas Fig~\ref{fig:Heatmap} emphasizes the ranks of these predicted drugs for the COVID-19 nodes. The majority of the corticosteroids we predict belong to the respiratory system (R) class, which has been the primary target for the coronaviruses, as reflected by the symptoms. However, COVID-19 has a multi-organ impact on the human body and is not limited to the respiratory system~\cite{Multiorgan}. Complications due to the cytokine storm with the effects of angiotensin converting enzyme (ACE) have led to cardiac arrest, kidney failure, and liver damage resulting in many deaths. For these reasons, we see drugs from various ATC classes are being considered for clinical trials. Next, we discuss in detail these pharmacological classifications of some of the predicted drugs. 

\textbf{Anti-inflammatory (AI) agents}: Inflammatory cytokine storms are prominently evident in COVID-19 positive patients and timely anti-inflammation treatment is required~\cite{AI_1}. Pneumonia caused by the coronavirus results in a huge amount of inflammatory cell infiltration leading to acute respiratory distress syndrome (ARDS), causing many deaths~\cite{AI_2,AI_3}. A wide range of anti-inflammatory treatments including glucocorticoids, non-steroidal anti-inflammatory drugs (NSAIDs), immunosuppressants, inflammatory cytokines antagonists like tumor necrosis factor (TNF) inhibitors, Janus kinase (JAK) inhibitors, and interleukin-1-receptor antagonist (IL-1RA) are being considered for COVID-19. Our model predicts in the top 10, steroids like \emph{Dexamethasone}, \emph{Hydrocortisone}, \emph{Methylprednisolone}; NSAIDs like \emph{Ibuprofen}, \emph{Aspirin (Acetylsalicylic acid)}; immunosuppressants like \emph{Sirolimus} and \emph{Methotrexate}; IL-1RA \emph{Anakinra}; CoX2 inhibitor \emph{Celecoxib}. These drugs are currently undergoing clinical trials. Some of the other corticosteroids like \emph{Betamethasone}, \emph{Prednisone}, and TNF inhibitor \emph{Certolizumab pegol} were also ranked high. 

\textbf{Antiviral and anti-parasitic agents}: \texttt{Dr-COVID} predicts nucleotide analogue antivirals like \emph{Acyclovir}, \emph{Valaciclovir}, \emph{Cidofovir}, and \emph{Entecavir} that have shown positive results in terminating the RNA synthesis catalyzed by polymerases of coronaviruses~\cite{Nucleotide}. \emph{Ivermectin} and \emph{Nitazoxanide} are used against many parasite infestations and are also known to have antiviral properties. \emph{Mebendazole} is another similar anti-parasitic drug that \texttt{Dr-COVID} ranked high. One of the recent reports shows that \emph{Ivermectin} is an effective inhibitor of the \emph{SARS-CoV-2} and many other positive single-stranded RNA viruses. A 5000-fold reduction in the virus titer within 48 hours in cell culture was obtained with a single treatment (5$\mu M$) of \emph{Ivermectin}~\cite{Ivermectin}. 

\textbf{Statins and ACE inhibitors/ beta-blockers/ calcium channel blockers}: Statins are lipid-lowering drugs that inhibit the cholesterol synthesis enzyme (also known as HMG-CoA reductase), which also has anti-inflammatory properties. There have been implications of lipid metabolism in the \emph{SARS-CoV-2} pathogenesis~\cite{Statin}, due to which there are reports on including statins in the line of treatment for COVID-19. \texttt{Dr-COVID} predicts \emph{Atorvastatin}, \emph{Simvastatin}, and \emph{Rosuvastatin}, where all the three drugs are currently in clinical trials. On the contrary, some studies show that statins tend to increase the cellular expression of ACE inhibitors~\cite{Statin_1}, to which the \emph{SARS-CoV2-spike} protein binds at the entry-level in humans~\cite{ACE_1}. Analyzing this issue, an observational study by Zang et al.~\cite{ACE_2} reported a reduced mortality rate in the patients treated with statins and no adverse effect was observed by adding an ACE inhibitor drug also to the line of treatment. These ACE inhibitors are cardiovascular drugs causing relaxation of blood vessels that are primarily used to treat high blood pressure and heart failure. Beta-adrenergic and calcium channel blockers are other similar functioning drugs that lower blood pressure, are also currently considered to treat COVID-19. \texttt{Dr-COVID} predicts \emph{Captopril} (ACE inhibitor), \emph{Atenolol} (beta-blocker) and \emph{Nifedipine} (calcium channel blocker), which are currently in clinical trials. Additionally, the list of predicted drugs includes \emph{Spironolactone} and \emph{Hydrochlorothiazide}, that help prevent our body from absorbing too much salt and eventually lowering the blood pressure and avoiding cardiac failure.

\textbf{Miscellaneous}: \texttt{Dr-COVID} also predicts some of the pre-discovered vaccines such as \emph{Rubella virus vaccine}, which is majorly considered for all the healthcare workers, the \emph{Yellow fever vaccine}~\cite{YFV}, and the \emph{Ebola zaire virus vaccine (rVSV-ZEBOV)}. Further, we also have  \emph{Mercaptopurine}, an antineoplastic agent that has been considered as a selective inhibitor of \emph{SARS-CoV}~\cite{Mercaptopurine} in the list of predicted drugs. Antidepressant \emph{Brexanolone} that is currently considered for patients on ventilator support due to ARDS, vasodilators \emph{Nitroglycerine} and \emph{Alprostadil}, nutritional supplements like \emph{Riboflavin} (Vitamin B\textsubscript{2})~\cite{Riboflavin}, \emph{Niacin}, \emph{Cholecalciferol} (Vitamin D\textsubscript{3}), and \emph{Iron} are some more top-ranked drugs. Interestingly, \emph{Ephedra sinica root}, a herb generally used to treat asthma and lung congestion, and an ingredient of lung cleansing and detoxifying decoction (LCDD), which is a widely used traditional Chinese medicine~\cite{TCM} is one of the drugs predicted in our list.

In essence, \texttt{Dr-COVID} predicts drugs for COVID-19 from different pharmacological classes like the corticosteroids, antivirals, antiparasitic, NSAIDs, and cardiovascular drugs, as the disease does not target particular anatomy and impacts multiple organs in the human body.

\section{Dataset} \label{sec:Dataset}
\begin{figure*}
        \centering
        \includegraphics[scale=0.75]{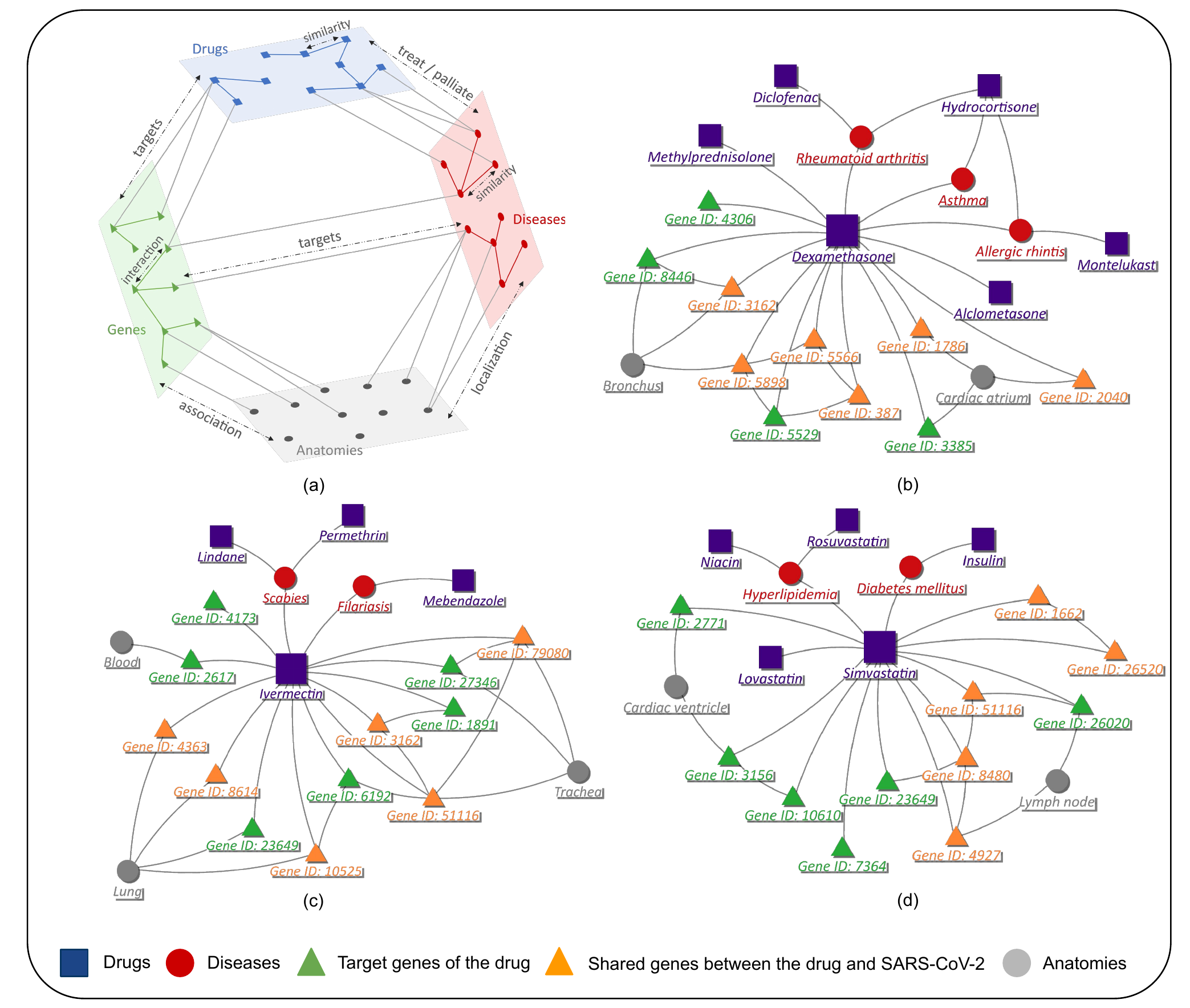}
       \caption{(a) Four-layered heterogeneous graph illustrating the inter-layer and intra-layer connections. (b), (c) and (d) Subgraph centered around the drugs \emph{Dexamethasone}, \emph{Ivermectin} and \emph{Simvastatin}, respectively.}
        \label{fig:DR_graph}
\end{figure*}

In this section, we describe the dataset that we use to train and test \texttt{Dr-COVID} for COVID-19 drug repurposing. We also describe how we model the data as a multilayer graph to capture the underlying complex interactions between different biological entities. We derive the required information from DRKG, which is a comprehensive biological knowledge graph relating genes, drugs, diseases, biological processes, side effects, and other eight more entities useful for computational pharmacological tasks like drug repurposing, drug discovery, and drug adverse effect prediction, to list a few. DRKG gathers all this information from six databases, namely, Drugbank~\cite{Drugbank}, Hetionet~\cite{Hetionet}, GNBR~\cite{GNBR}, STRING~\cite{STRING}, IntAct~\cite{IntAct}, and DGIdb~\cite{DGIdb}. From DRKG, we consider four entities that are relevant to the drug repurposing task. The four entities are drugs (e.g., \emph{Dexamethasone}, \emph{Sirolimus}), diseases (e.g., \emph{Scabies}, \emph{Asthma}), anatomies (e.g., \emph{Bronchus}, \emph{Trachea}), and genes (e.g., \emph{Gene~ID:~8446}, \emph{Gene~ID:~5529}). All the genes are referred with their respective Entrez IDs throughout the paper. We extract the details about these entities specifically from the Drugbank, Hetionet, and GNBR databases. We form a four-layered heterogeneous graph with these four entities in each layer as illustrated in Fig~\ref{fig:DR_graph}a. The four-layered graph is composed of 8070 drugs, 4166 diseases, 29848 genes, 400 anatomies, and a total of 1,417,624 links, which include all the inter-layer and intra-layer connections. Next, we discuss the interactome that we consider for drug repurposing.

\textbf{Interactome}:  There are inter-layered connections between the four layers and some have intra-layered connections. The inter-layered connections are of different types. The drug-disease links indicate treatment or palliation, i.e., a drug treats or has a relieving effect on a disease. For example, interaction between \emph{Ivermectin-Scabies} (as seen in Fig~\ref{fig:DR_graph}b)  and \emph{Simvastatin-Hyperlipidemia} (as seen in Fig~\ref{fig:DR_graph}d) are of type treatment, whereas \emph{Atropine-Parkinson's disease} and \emph{Diclofenac-Osteoarthritis} are of type palliation. The drug-gene and disease-gene links are the direct gene targets of the compound and the disease, respectively. \emph{Gene~ID:~4306}, \emph{Gene~ID:~387}, \emph{Gene~ID:~1786} are some of the targets of the drug \emph{Dexamethasone} (see Fig~\ref{fig:DR_graph}b) and \emph{Gene~ID:~5509}, \emph{Gene~ID:~859} are target genes of the disease \emph{Malaria}. Some of the genes targeted by the drug (e.g., \emph{Dexamthasone}, \emph{Ivermectin}, \emph{Simvastatin}) as well as by the \emph{SARS-CoV-2} virus (referred to as shared genes) are shown in Fig~\ref{fig:DR_graph} (b,c and d). These common gene targets between a drug and a disease are one of the reasons for the drug to be a potential repurposing candidate against the disease. The disease-anatomy and gene-anatomy connections indicate how the diseases affect the anatomies and interactions between the genes and anatomies. For example, \emph{Gene~ID:~2771} and \emph{Gene~ID:~3156} belong to the \emph{cardiac ventricle} anatomy (see Fig~\ref{fig:DR_graph}d); disease \emph{Schizophrenia} affects multiple anatomies like the \emph{central nervous system (CNS)} and \emph{optic tract}.

There are also intra-layered connections. The drug-drug and disease-disease connections show the similarity between a pair of drugs and diseases, respectively. The gene-gene links describe the interaction between genes (e.g., epistasis, complementation) and form the whole gene interactome network. This comprehensive gene network serves as a backbone for our model, wherein we predict the unknown links between drugs and new diseases like COVID-19 as they are connected through genes and anatomies. The anatomy information helps in drug predictions by focusing on the local interactions of genes related to the same anatomy as the genes targeted by the disease. Some examples of the intra-layered connections are \emph{Simvastatin}-\emph{Lovastatin} and \emph{Gene~ID:~23649}-\emph{Gene~ID:~8480} as seen in Fig~\ref{fig:DR_graph}d. While all these interactions reveal the true relations between the entities, we also randomly sample the no-drug-disease links, which give us negative control in the learning process. For example, there is no link between \emph{Simvastatin-Scabies}, i.e., \emph{Simvastatin} is not known to treat or suppress the effects of \emph{Scabies}. Including such negative control in the training process makes our model accurate and reliable.

\textbf{HCoV interactome network}: To specialize the drug repurposing model \texttt{Dr-COVID} for COVID-19, as discussed before,  we consider the four known HCoVs, namely, \emph{SARS-CoV}, \emph{MERS-CoV}, \emph{CoV-229E} and \emph{CoV-NL63}, and two non-human CoVs namely \emph{MHV}, and \emph{IBV}. We consider interactions of these disease nodes with human genes. There are 129 links between these six disease nodes and the gene nodes~\cite{DRKG}. In addition, we consider all the 27 \emph{SARS-CoV-2} proteins (including the structured proteins, nsp, and orf) and their 332 links connecting the target human genes as given by Gordon et al.~\cite{Gordon}. In other words, there are only disease-gene interactions available for these COVID-19 nodes. With this available information, we train \texttt{Dr-COVID} to predict possible drug connections for these COVID-19 nodes.

\section{Methods and Models} \label{sec:Methods}
In the last few years, deep learning has gained significant attention from a variety of scientific disciplines due to its extraordinary successes in solving many challenging tasks like data cleansing, mining, and classification, mainly for images, speech, or text datasets.  However, in many applications, the structure underlying data is not always Euclidean. Some examples include social networks, transportation networks, brain networks, sensor networks, chemical molecules, protein-protein interactions, meshed surfaces in computer graphics, and the drug repurposing network, as discussed above, to list a few. For these applications, more recently, deep learning for graph-structured data, also known as geometric deep learning (GDL)~\cite{Bronstein}, is receiving steady research attention. GDL aims at building neural network architectures known as graph neural network (GNNs) to learn from graph-structured data. GDL models are used to learn low-dimensional graph representations or node embeddings by taking into account the nodal connectivity information. These embeddings are then used to solve many graph analysis tasks like node classification, graph classification, and link prediction, to list a few. GNN architectures are developed using concepts from spectral graph theory and generalize the traditional convolution operation in the convolutional neural network (CNN) to the graph setting. In this section, we describe the proposed \texttt{Dr-COVID} architecture for COVID-19 drug repurposing and describe numerical experiments performed to evaluate our model.

Consider an undirected graph  $\cal{G} = (\cal{V},\cal{E})$ with a set of vertices $\cal{V} =$ \{$v_1, v_2, \cdots, v_N$\} and edges $e_{ij} \in \cal{E}$ denoting a connection between nodes $v_i$ and $v_j$. We represent a graph $\cal{G}$ using the adjacency matrix $\mathbf{A} \in \mathbb{R}^{N\times N}$, where the $(i,j)$th entry of $\mathbf{A}$ denoted by $a_{ij}$ is $1$ if there exists an edge between nodes $v_i$ and $v_j$, and $zero$ otherwise. To account for the non-uniformity in the degrees of the nodes, we use the normalized adjacency matrix denoted by $\tilde{\mathbf{A}}=\mathbf{D}^{-\frac{1}{2}}\mathbf{A}\mathbf{D}^{-\frac{1}{2}}$, where $\mathbf{D} \in \mathbb{R}^{N\times N}$ is the diagonal degree matrix. Each node in the graph is associated with its own feature vector (referred to as input feature). Let us denote the input feature of node $i$ by $\mathbf{x}_{i}^{(0)} \in \mathbb{R}^d$, which contains key information or attributes of that node (e.g., individual drug side effects). Let $\mathbf{X}^{(0)} \in \mathbb{R}^{N\times d}$ be the input feature matrix associated with the $N$ nodes in the graph $\cal{G}$ obtained by stacking the input features of all the nodes in $\cal{G}$. The new embeddings for a node is generated by combining information from its neighboring nodes (e.g., diseases or genes) to account for the local interactions. This process of combining information and generating new representations for a node is done by a single GNN block. If we stack $K$ such blocks, we can incorporate information for a node from its $K$-hop neighbors (e.g., in Fig~\ref{fig:DR_graph}c, the drug \emph{Ivermectin} is a $2$-hop neighbor of the anatomy \emph{Lung} and is connected via \emph{Gene~ID:~8614}). Mathematically, this operation can be represented as
\begin{eqnarray}
\label{eq:GCN1}
	\mathbf{X}^{(k+1)} = g_k(\bar{\mathbf{A}}\mathbf{X}^{(k)}\mathbf{W}_k),
\end{eqnarray}
where $\mathbf{X}^{(k)} \in \mathbb{R}^{N\times d_{k}}$ represents the $k$th layer embedding matrix and $d_k$ is the embedding dimension in the $k$th layer. Here, $\bar{\mathbf{A}} = \mathbf{I}+\tilde{\mathbf{A}}$, where the identity matrix $\mathbf{I}\in \mathbb{R}^{N\times N}$, is added to account for the self-node embeddings, $\textbf{W}_{k} \in \mathbb{R}^{d_k\times d_{k+1}}$ is the learnable transformation matrix, and $g_k(\cdot)$ is the activation function in the $k$th layer. There exist several GNN variants such as graph convolutional networks (\texttt{GCN})~\cite{GCN}, \texttt{GraphSAGE}~\cite{GraphSAGE}, graph attention networks (\texttt{GAT})~\cite{GAT} and scalable inception graph neural network (\texttt{SIGN})~\cite{SIGN}, to name a few. \texttt{GCN} is a vanilla flavored GNN based on Eq~\eqref{eq:GCN1}. \texttt{GAT} gives individual attention to the neighboring nodes instead of treating every node equally. To address the issue of scalability, \texttt{GraphSAGE} uses a neighbor sampling method, wherein instead of taking the entire neighborhood, we randomly sample a subset of neighbor nodes. \texttt{SIGN} takes a different approach to solve the scalability issue and introduce a parallel architecture. The proposed \texttt{Dr-COVID} architecture is based on the \texttt{SIGN} approach due to its computational advantages. The predicted list of drugs from other GNNs are available in our repository. Next, we describe the proposed \texttt{Dr-COVID} architecture.

\subsection{\textbf{\texttt{Dr-COVID} architecture}}
\begin{figure*}
    \centering
    \includegraphics[angle=270,scale=0.155]{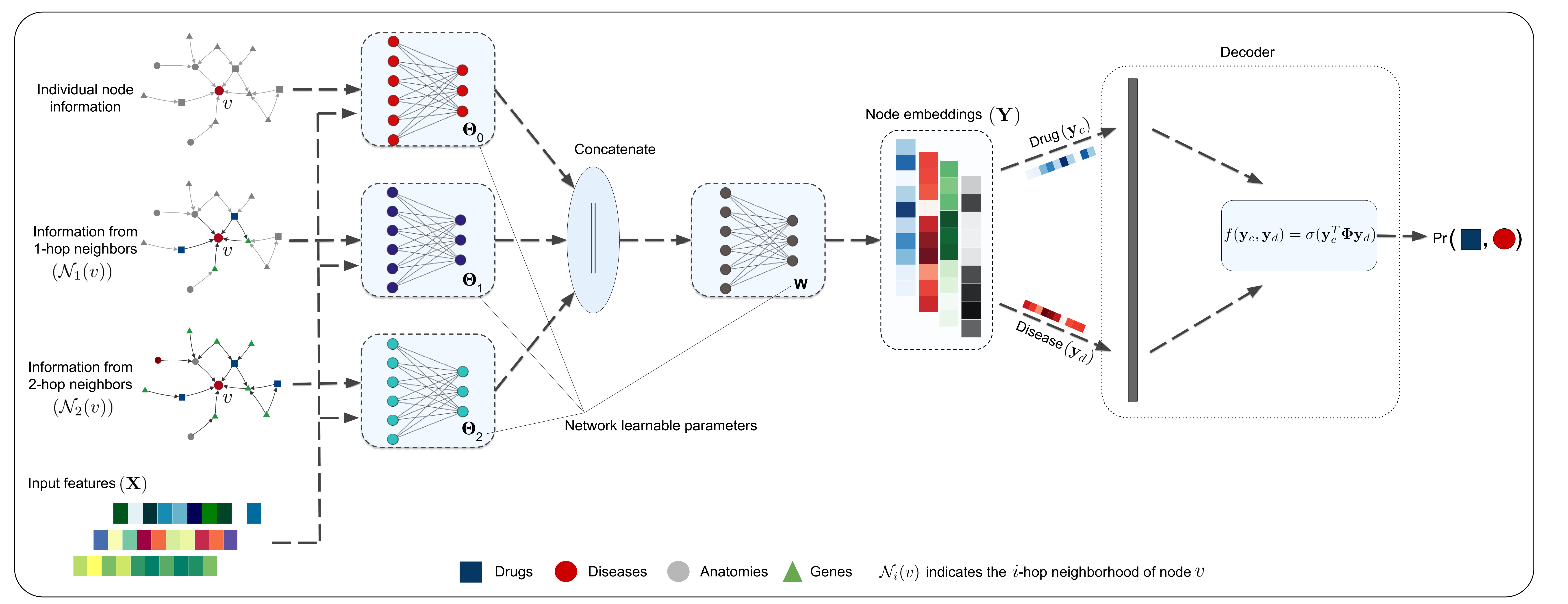}
     \caption{\texttt{Dr-CoV} architecture.}
    \label{fig:architecture}
\end{figure*}

The proposed GNN architecture for \emph{SARS-CoV-2} drug repurposing has two main components, namely, the encoder and decoder. The encoder based on the \texttt{SIGN} architecture generates the node embeddings of all the nodes in the four-layer graph. The decoder scores a drug-disease pair based on the embeddings. The encoder and decoder networks are trained in an end-to-end manner. Next, we describe these two components of the \texttt{Dr-COVID} architecture, which is illustrated in Fig~\ref{fig:architecture}.

\textbf{Encoder}: The \texttt{Dr-COVID} encoder is based on the \texttt{SIGN} architecture~\cite{SIGN}, which provides low-dimensional node embeddings based on the input features and nodal connectivity information. Recall that the matrix $\mathbf{A}$ is the adjacency matrix of the four-layered graph $\cal{G}$ and $\tilde{\mathbf{A}}$ is the normalized adjacency. \texttt{SIGN} uses linear diffusion operators represented using matrices $\mathbf{F}_{r}$, $r=1,2,\cdots$, to perform message passing and aggregate local information in the graph. By choosing $\mathbf{F}_r = \tilde{\mathbf{A}}^r$ we can incorporate information for node $v$ from its $r$-hop neighbors. Here, $\tilde{\mathbf{A}}^r$ denotes the $r$th matrix power. To start the information exchange between the nodes, we assume that each node has its own $d$ dimensional feature, which we collect in the matrix $\mathbf{X} \in \mathbb{R}^{N\times d}$ to obtain the complete input feature matrix associated with the nodes of $\cal{G}$. We can then represent the encoder as 
\begin{eqnarray}
\label{eq:SIGN1}
	\mathbf{Z} = \sigma_1\left\{\left[\mathbf{X}\boldsymbol{\Theta}_0\, \|\,\mathbf{F}_1\mathbf{X}\boldsymbol{\Theta}_1\|\,\cdots\|\,\mathbf{F}_r\mathbf{X}\boldsymbol{\Theta}_r\right]\right\} \quad \text{and} \quad \mathbf{Y} = \sigma_2\left\{\mathbf{ZW}\right\},
\end{eqnarray}
where $\textbf{Y}$ is the final node embedding matrix for the nodes in the graph $\cal{G}$, and \{$\boldsymbol{\Theta}_0,\cdots,\boldsymbol{\Theta}_r,\mathbf{W}$\} are the learnable parameters. Here,~$\|$ represents concatenation and $\sigma_1\{\cdot\}$ and $\sigma_2\{\cdot\}$ are the nonlinear tanh and leaky rectified linear unit (leaky ReLU) activation functions, respectively. The matrix $\mathbf{F}_r\mathbf{X} = \mathbf{D}^{-\frac{1}{2}}\mathbf{A}^r\mathbf{D}^{-\frac{1}{2}}\mathbf{X}$ captures information about the local interactions over $r$-hop neighbors. Fig~\ref{fig:architecture} shows the encoder architecture. The main benefit of using \texttt{SIGN} over other sequential models (e.g., \texttt{GCN}, \texttt{GAT}, \texttt{GraphSAGE}) is that the matrix product $\textbf{F}_r\textbf{X}$ is independent of the learnable parameters $\boldsymbol{\Theta}_r$. Thus, this matrix product can be pre-computed before training the neural network model. Doing so reduces the computational complexity without compromising the performance.

In our setting, we choose $r=2$, i.e., the low-dimensional node embeddings have information from 2-hop neighbors. Choosing $r \geq 3$ is not useful for drug repurposing, as we aim to capture the local information of the drug targets such that a drug node embedding should retain information about its target genes and the shared genes in its vicinity. For example, the $1$-hop neighbors of \emph{Dexamethasone} as shown in Fig~\ref{fig:DR_graph}b, are the diseases it treats (e.g., \emph{Asthma}), and the drugs similar to \emph{Dexamethasone} (e.g., \emph{Methylprednisolone}) and its target genes (e.g., \emph{Gene~ID:~8446}, \emph{Gene~ID:~387}). The $2$-hop neighbors are the anatomies of the target genes (e.g., \emph{Bronchus}) of \emph{Dexamethasone}, and the drugs that have similar effects on the diseases (e.g., \emph{Hydrocortisone} and \emph{Dexamethasone} have similar effects on \emph{Asthma}). It is essential for the embedding related to \emph{Dexamethasone} to retain this local information for the drug repurposing task, and not much benefit is obtained by propagating more deeper in the network. 

\textbf{Decoder}: For drug repurposing, we propose a score function that takes as input the embeddings of the drugs and diseases and outputs a score based on which we decide if a certain drug treats the disease. Fig~\ref{fig:architecture} illustrates the proposed decoder. The columns of the embedding matrix $\textbf{Y}$, contains the embeddings of all the nodes in the four-layer graph, including the embeddings of the disease and drug nodes. Let us denote the embeddings of the $i$th drug as ${\bf y}_{c_i} \in \mathbb{R}^l$ and the embeddings of the $j$th disease as ${\bf y}_{d_j} \in \mathbb{R}^l$. The proposed scoring function $f(\cdot)$ to infer whether drug $c_i$ is a promising treatment for disease $d_j$ is defined as
\begin{eqnarray}
\label{eq:Scorer}
	s_{ij} = f({\bf y}_{c_i},{\bf y}_{d_j}) = \sigma \left\{{\bf y}^T_{c_i}{\boldsymbol \Phi}{\bf y}_{d_j}\right\},
\end{eqnarray}
where $\sigma \{\cdot\}$ is the nonlinear sigmoid activation function and ${\boldsymbol \Phi} \in \mathbb{R}^{l\times l}$ is a learnable co-efficient matrix. We interpret $s_{ij}$ as the probability that a link exists between drug $c_i$ and disease $d_j$.  The term ${\bf y}^T_{c_i}{\boldsymbol \Phi}{\bf y}_{d_j}$ can be interpreted as a measure of correlation (induced by ${\boldsymbol \Phi}$) between the disease and drug node embeddings. We use $d=400$ and $l=250$ in our implementation. The model is trained in a mini-batch setting in an end-to-end fashion using stochastic gradient descent to minimize the weighted cross entropy loss, where the loss function for the sample corresponding to the drug-disease pair $(i,j)$ is given by
\begin{eqnarray}
\label{eq:Loss}
	\ell(s_{ij},z_{ij}) = wz_{ij}\left(\log\left(\frac{1}{\sigma (s_{ij})}\right)\right) + \left(1-z_{ij}\right)\log\left(\frac{1}{1-\sigma (s_{ij})}\right),
\end{eqnarray}
where $z_{ij}$ is the known training label associated with score $s_{ij}$ for the drug-disease pair, $z_{ij}=1$ indicates that drug $i$ treats disease $j$ and otherwise when $z_{ij}=0$. Here, $w$ is the weight on the positive samples that we choose to account for the class imbalance. As discussed in the Dataset Section, we include the no-drug-disease links as negative control while training our model. The number of no-drug-disease links is almost thirty times the number of positive samples. To handle this class disparity, we explicitly use a weight $w>0$ on the positive samples. %Next, we discuss the choice of $w$ and other hyper-parameters (e.g., batch size), and evaluate \texttt{Dr-CoV}.

\subsection{Model evaluation}
In this subsection, we evaluate \texttt{Dr-COVID} and discuss the choice of various hyper-parameters. The drug repurposing via link prediction can be viewed as a binary classification problem, wherein a positive class represents the existence of a link between the input drug and disease, and otherwise for a negative class. We have 6113 positive samples (drug-disease links) in our dataset. To account for the negative class samples, we randomly choose 200,000 no-drug-disease links (i.e., there is no link between these drugs and diseases). These links are then divided into the training and testing set with a $90\%-10\%$ split. To use mini-batch stochastic gradient descent, we group the training set in batches of size 512 and train them for 20 epochs. Due to the significant class imbalance, we oversample the drug-disease links while creating batches, thus maintaining the class ratio (ratio of the number of negative samples to the number of positive samples) of $1.5$ in each batch. The weight $w$ on the positives samples (mentioned in Eq~\eqref{eq:Loss}), is also chosen to be the class imbalance ratio of each batch, i.e., we fix $w$ to be $1.5$.

We perform experiments on three sequential GNN encoder architectures, namely, \texttt{GCN}~\cite{GCN}, \texttt{GraphSAGE}~\cite{GraphSAGE}, and \texttt{GAT}~\cite{GAT} for the drug repurposing task, which we treat as a link prediction problem, and compare with the proposed \texttt{Dr-COVID} architecture. Specifically, the \texttt{SIGN} encoder in \texttt{Dr-COVID} is replaced with \texttt{GCN}, \texttt{GraphSAGE}, and \texttt{GAT} to evaluate the model performance. Two blocks of these sequential models are stacked to maintain the consistency with $r=2$ of the \texttt{Dr-COVID} architecture. We evaluate these models on the test set, which are known treatments for diseases that are not shown to the model while training. The model is evaluated based on two performance measures. Firstly, we report the ability to classify the links correctly, i.e., to predict the known treatments correctly for diseases in the test set. This is measured through the receiver operating characteristic (ROC) curve of the true positive rate (TPR) versus the false positive rates (FPR). Next, using the list of predicted drugs for the diseases in the test set, we report that model's ability to rank the actual treatment drug as high as possible (the ranking is obtained by ordering the scores in Eq~\eqref{eq:Scorer}). 

\begin{figure}
    \centering
    \includegraphics[scale=0.5]{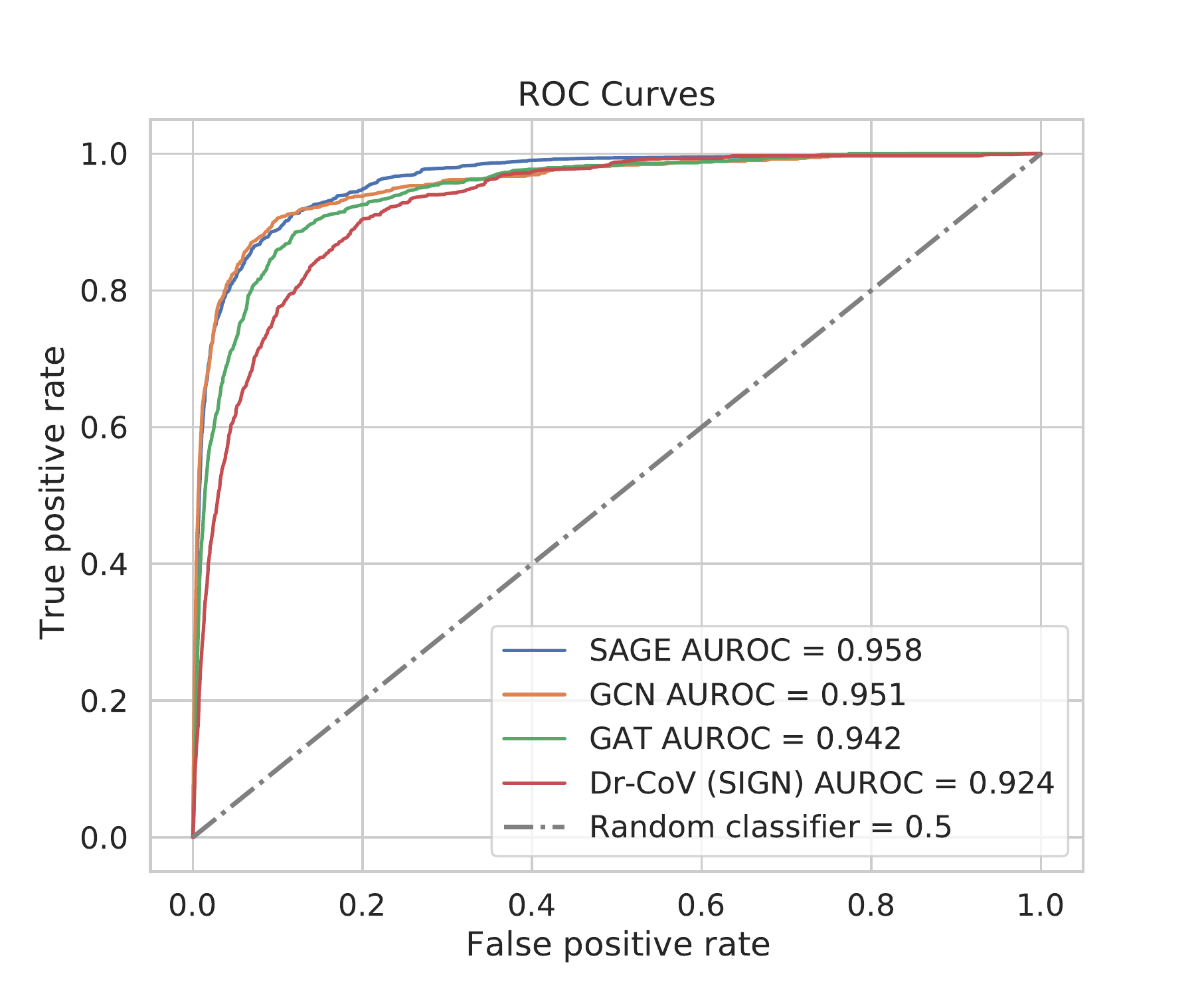}
    \caption{ROC Curves.}
    \label{fig:ROC}
\end{figure}

ROC curves show the performance of a binary classification model by varying the threshold values used to classify the positive samples, which eventually change the TPR and FPR. Fig~\ref{fig:ROC} shows the ROC curves of different GNN models. The area under ROC (AUROC), which lies in the interval [0,1] indicates the separation ability of a binary classifier, where 1 indicates the best performance, 0.5 means that the model is unable to discriminate between the classes and 0 indicates a completely opposite behavior. We can see from Fig~\ref{fig:ROC} that all the models have very similar AUROC values.

We also evaluate \texttt{Dr-COVID} in terms of ranks of the actual treatment drug in the predicted list for a disease from the testing set, where the rank is computed by rank ordering the scores as before. In addition, we compute the network proximity scores~\cite{DR_M_1} and rank order the drugs based on these scores to compare with other GNN encoder models. These network proximity scores are a measure of the shortest distance between drugs and diseases. They are computed as 
\begin{eqnarray}
\label{eq:dct}
	P_{ij} = \frac{1}{|\cal{C}|+|\cal{T}|}\left(\sum_{p\in \cal{C}}\min_{q\in \cal{T}}d(p,q) + \sum_{q \in \cal{T}}\min_{p\in \cal{C}}d(p,q)\right),
\end{eqnarray}
where $P_{ij}$ is a proximity score of drug $c_i$ and disease $d_j$. Here, $\cal{C}$ is the set of target genes of $c_i$, $\cal{T}$ is the set of target genes of $d_j$, and $d(p,q)$ is the shortest distance between a gene $p \in \cal{C}$ and a gene $q \in \cal{T}$ in the gene interactome. We convert these into Z-scores using the permutation test as
\begin{eqnarray}
\label{eq:Z_score}
	Z_{ij} = \frac{P_{ij}-\mu}{\omega},
\end{eqnarray}
where $\mu$ is the mean proximity score of $c_i$ and $d_j$, which we compute by randomly selecting subsets of genes with the same degree distribution as that of $\cal{C}$ and $\cal{T}$ from the gene interactome, and $\omega$ is the standard deviation of the scores generated in the permutation test. Table~\ref{tab:rank_table} gives the rankings, which clearly show that the \texttt{Dr-COVID} results in better ranks on the unseen diseases than the other GNN variants. Also, compared to the network proximity measure, which is solely based on the gene interactome, \texttt{Dr-COVID} performs better. We choose these drug-disease pairs for evaluation as these links are not shown during the training.  It is evident that the diseases on which we evaluate are not confined to single anatomy (e.g., \emph{rectal neoplasms} are associated to the \emph{rectum} anatomy, whereas \emph{pulmonary fibrosis} is a \emph{lung} disease), nor do they require a similar family of drugs for their treatment (e.g., \emph{Fluorouracil} is an antineoplastic drug, and \emph{Prednisone} is an anti-inflammatory corticosteroid). Thus, showcasing our model's unbiased nature. For a majority of the diseases in the test set \texttt{Dr-COVID} ranks the treatment drug in top 10 (as seen in Table~\ref{tab:rank_table}). In the case of \emph{Leukemia} (blood cancer), other antineoplastic drugs like \emph{Hydroxyurea} and \emph{Methotrexate} are ranked high (in top $10$) and its known treatment drug \emph{Azacitidine} is ranked $17$. We give more importance to the ranking parameter as any drug predictor requires classifying and ranking the correct drugs as high as possible. Considering this AUROC-ranking trade-off we can see that \texttt{Dr-COVID} with \texttt{SIGN} encoder performs the best.
\begin{table*}[!htbp]
    \centering
    \resizebox{\columnwidth}{!}{
    \begin{tabular}{*7c}
        \hline
        \hline
        Disease &  Treatment drug & \multicolumn{5}{c}{Ranks}\\
        {}   & {}  & \texttt{Dr-CoV (SIGN)}   & \texttt{SAGE} & \texttt{GCN} & \texttt{GAT} & Network proximity\\
        \hline
        \emph{Encephalitis} & \emph{Acyclovir} & \textbf{10} & 35 & 35 & 295 & 5462\\
        \emph{Rectal neoplasms} & \emph{Fluorouracil} & \textbf{9} & 421 & 16 & 231 & 2831\\
        \emph{Pulmonary fibrosis} & \emph{Prednisone} & 5 & \textbf{3} & 10 & 9 & 2072\\
        \emph{Atrioventricular block} & \emph{Atropine} & \textbf{6} & 79 & 8 & 14 & 4453\\
        \emph{Pellagra} & \emph{Niacin} & \textbf{2} & 56 & 497 & 484 & Not computable \\
        \emph{Colic} & \emph{Hyoscyamine} & \textbf{1} & \textbf{1} & 501 & 205 & Not computable \\ 
        \emph{Leukemia} & \emph{Azacitidine} & \textbf{17} & 120 & 31 & 332 & 377 \\
        \hline
        \hline
    \end{tabular}
    }
    \caption{{\bf Ranking Table.} The Table gives the ranking performance of \texttt{Dr-COVID} compared with other GNN variants and the network proximity measures. There are no associated genes with some of the disease in our database, which makes it impossible to compute the Z scores. These are indicated as ``Not computable". The best results are highlighted in bold font.}
    \label{tab:rank_table}
\end{table*}

\textbf{COVID-19 analysis}: We perform a similar analysis and identify potential candidate drugs for \emph{SARS-CoV-2}. For all the COVID-19 nodes in our dataset comprising 27 proteins (structured, nsp and orf), \emph{SARS-CoV}, \emph{IBV}, \emph{MERS-CoV}, \emph{CoV-229E}, \emph{CoV-NL63}, and \emph{MHV}, we individually predict the drugs for all these 33 entities. Each protein in \emph{SARS-CoV-2} targets a different set of genes in humans, so we give individual predictions. We then pick the top 10 drugs from all the predicted drugs and list 150 candidate repurposed drugs for COVID-19. Out of these 150 drugs, 46 are currently in clinical trials. Our predictions have a mixture of antivirals, antineoplastic, corticosteroids, monoclonal antibodies (mAb), non-steroidal anti-inflammatory drugs (NSAIDs), ACE inhibitors, and statin family of drugs, and some of the vaccines discovered previously for other diseases. Refer to the Results Section for a detailed discussion on the analysis of the predicted drugs for COVID-19.

\section{Conclusions} \label{sec:conclusions}
In this work, we presented a generalized drug repurposing model, called \texttt{Dr-COVID} for novel human diseases. We constructed a biological network of drugs, diseases, genes, and anatomies and formulated the drug repurposing task as a link prediction problem. We proposed a graph neural network model, which was then trained to predict drugs for new diseases. \texttt{Dr-COVID} predicted 150 potential drugs for COVID-19, of which 46 drugs are currently in clinical trials. The considered GNN model is computationally efficient and better ranks known treatment drugs for diseases than the other GNN variants and non-deep methods like the network proximity approaches. This work can be extended along several directions. Considering the availability of substantial biological data, the inclusion of information like individual side effects of drugs, the molecular structure of the drugs, etc., may further improve the predictions. Considering the comorbidities of a patient would help us analyze the biological process and gene interactions in the body specific to an individual and accordingly prescribe the line of treatment. Predicting a synergistic combination of drugs for a disease would be another area of interest where graph neural networks can be beneficial.

\section*{Acknowledgements} \label{sec:acknowledgements}
S.P.~Chepuri is supported in part by the Pratiskha Trust Young Investigator Award,  Indian Institute of Science, Bangalore, and the SERB grant SRG/2019/000619, and S.~Doshi is supported by the Robert Bosch Center for Cyber Physical Systems, Indian Institute of Science, Bangalore, Student Research Grant 2020-M-11 . 

The authors thank the Deep Graph Learning team for compiling DRKG and making the data public at \url{https://github.com/gnn4dr/DRKG}.

\section*{Data and Code availability} \label{sec:code}

All the implementation and the data required to reproduce the results in the paper are available at \url{https://github.com/siddhant-doshi/Dr-COVID}. 

%\input{sections/ref.tex}

%\appendix*
%\input{sections/appendix.tex}

\begin{comment}

\newpage

\end{comment}


%apsrev4-2.bst 2019-01-14 (MD) hand-edited version of apsrev4-1.bst
%Control: key (0)
%Control: author (8) initials jnrlst
%Control: editor formatted (1) identically to author
%Control: production of article title (0) allowed
%Control: page (0) single
%Control: year (1) truncated
%Control: production of eprint (0) enabled
\begin{thebibliography}{0}%
\makeatletter
\providecommand \@ifxundefined [1]{%
 \@ifx{#1\undefined}
}%
\providecommand \@ifnum [1]{%
 \ifnum #1\expandafter \@firstoftwo
 \else \expandafter \@secondoftwo
 \fi
}%
\providecommand \@ifx [1]{%
 \ifx #1\expandafter \@firstoftwo
 \else \expandafter \@secondoftwo
 \fi
}%
\providecommand \natexlab [1]{#1}%
\providecommand \enquote  [1]{``#1''}%
\providecommand \bibnamefont  [1]{#1}%
\providecommand \bibfnamefont [1]{#1}%
\providecommand \citenamefont [1]{#1}%
\providecommand \href@noop [0]{\@secondoftwo}%
\providecommand \href [0]{\begingroup \@sanitize@url \@href}%
\providecommand \@href[1]{\@@startlink{#1}\@@href}%
\providecommand \@@href[1]{\endgroup#1\@@endlink}%
\providecommand \@sanitize@url [0]{\catcode `\\12\catcode `\$12\catcode
  `\&12\catcode `\#12\catcode `\^12\catcode `\_12\catcode `\%12\relax}%
\providecommand \@@startlink[1]{}%
\providecommand \@@endlink[0]{}%
\providecommand \url  [0]{\begingroup\@sanitize@url \@url }%
\providecommand \@url [1]{\endgroup\@href {#1}{\urlprefix }}%
\providecommand \urlprefix  [0]{URL }%
\providecommand \Eprint [0]{\href }%
\providecommand \doibase [0]{https://doi.org/}%
\providecommand \selectlanguage [0]{\@gobble}%
\providecommand \bibinfo  [0]{\@secondoftwo}%
\providecommand \bibfield  [0]{\@secondoftwo}%
\providecommand \translation [1]{[#1]}%
\providecommand \BibitemOpen [0]{}%
\providecommand \bibitemStop [0]{}%
\providecommand \bibitemNoStop [0]{.\EOS\space}%
\providecommand \EOS [0]{\spacefactor3000\relax}%
\providecommand \BibitemShut  [1]{\csname bibitem#1\endcsname}%
\let\auto@bib@innerbib\@empty
%</preamble>
\end{thebibliography}%


\begin{thebibliography}{10}

\bibitem{economy}
World Bank. Global Economic Prospects, June 2020; 2020b.
\newblock \url{http://hdl.handle.net/10986/33748 }.

\bibitem{CoV_1}
{Zumla A, Chan JF, Azhar EI, Hui DS, Yuen KY.}
\newblock {Coronaviruses—drug discovery and therapeutic options}.
\newblock Nat Rev Drug Discov. {2016};{15}({5}):{327--347}.

\bibitem{CoV_2}
{Paules CI, Marston HD, Fauci AS.}
\newblock {Coronavirus infections—more than just the common cold}.
\newblock JAMA. {2020};{323}({8}):{707--708}.

\bibitem{CoV_4}
{Lu H, Stratton CW, Tang YW.}
\newblock {Outbreak of pneumonia of unknown etiology in Wuhan, China: The
  mystery and the miracle}.
\newblock J Med Virol. {2020};{92}({4}):{401--402}.

\bibitem{CoV_3}
{Chen N, Zhou M, Dong X, Qu J, Gong F, Han Y, et al.}
\newblock {Epidemiological and clinical characteristics of 99 cases of 2019 novel coronavirus pneumonia in Wuhan, China: a descriptive study}.
\newblock Lancet. {2020};{395}({10223}):{507--513}.

\bibitem{CoV_5}
{Sohrabi C, Alsafi Z, O’Neill N, Khan M, Kerwan A, Al-Jabir A, et al.}
\newblock {World Health Organization declares global emergency: A review of the
  2019 novel coronavirus (COVID-19)}.
\newblock Int J Surg. {2020};{76}:{71--76}.

\bibitem{DR}
{Pushpakom S, Iorio F, Eyers PA, Escott KJ, Hopper S, Wells A, et al.}
\newblock {Drug repurposing: progress, challenges and recommendations}.
\newblock Nat Rev Drug Discov. {2018};{18}({1}):{41--58}.

\bibitem{CB_data}
{Zitnik M, Nguyen F, Wang B, Leskovec J, Goldenberg A, Hoffman M.}
\newblock {Machine learning for integrating data in biology and medicine: Principles, practice, and opportunities}.
\newblock Inf Fusion. {2019};{50}:{71--91}.

\bibitem{DDI_ML}
{Cheng F, Zhao Z.}
\newblock {Machine learning-based prediction of drug–drug interactions by integrating drug phenotypic, therapeutic, chemical, and genomic properties}.
\newblock J Am Med Inform Assoc. {2014};{21}({e2}):{e278--e286}.

\bibitem{DDI_DL_1}
{Yi Z, Li S, Yu J, Tan Y, Wu Q, Yuan H, et al.}
\newblock {Drug-drug interaction extraction via recurrent neural network with multiple attention layers}.
\newblock {International Conference on Advanced Data Mining and Applications}. {2017};{10604}:{554--566}.

\bibitem{DDI_DL_2}
{Deng Y, Xu X, Qiu Y, Xia J, Zhang W, Liu S.}
\newblock {A multimodal deep learning framework for predicting drug-drug interaction events}.
\newblock {Bioinformatics}. {2020};{36}({15}):{4316--4322}.

\bibitem{Decagon}
{Zitnik M, Agrawal M, Leskovec J.}
\newblock {Modeling polypharmacy side effects with graph convolutional networks}.
\newblock {Bioinformatics}. {2018};{34}({13}):{i457--i466}.

\bibitem{DR_M_1}
{Cheng F, Desai RJ, Handy DE, Wang R, Schneeweiss S, Bar\'{a}basi AL, et al.}
\newblock {Network-based approach to prediction and population-based validation of in silico drug repurposing}.
\newblock Nat Commun. {2018};{9}({1}):{1--12}.

\bibitem{DR_M_2}
{Guney E, Menche J, Vidal M, Bar\'{a}basi AL.}
\newblock {Network-based in silico drug efficacy screening}.
\newblock Nat Commun. {2016};{7}({1}):{1--13}.

\bibitem{DR_M_3}
{Zhou Y, Hou Y, Shen J, Huang Y, Martin W, Cheng F.}
\newblock {Network-based drug repurposing for novel coronavirus 2019-nCoV/SARS-CoV-2}.
\newblock Cell Discov. {2020};{6}({1}):{1--18}.

\bibitem{DR_M_4}
{Cheng F, Lu W, Liu C, Fang J, Hou Y, Handy DE, et al.}
\newblock {A genome-wide positioning systems network algorithm for in silico drug repurposing}.
\newblock Nat Commun. {2019};{10}({1}):{1--14}.

\bibitem{DR_M_5}
{Gramatica R, Di Matteo T, Giorgetti S, Barbiani M, Bevec D, Aste T.}
\newblock {Graph theory enables drug repurposing–how a mathematical model can drive the discovery of hidden mechanisms of action}.
\newblock {PLOS One}. {2014};{9}({1}):{e84912}.

\bibitem{Network_medicine}
{Gysi DM, Valle \'{I}D, Zitnik M, Ameli A, Gan X, Varol O, et al.}
\newblock {Network medicine framework for identifying drug repurposing opportunities for COVID-19}
\newblock arXiv:1403.3301 [Preprint]. 2020 [cited 2020 Nov 20]. Available from: 
\url{https://arxiv.org/abs/2004.07229}.

\bibitem{Few_shot}
{Ioannidis VN, Zheng D, Karypis G. }
\newblock {Few-shot link prediction via graph neural networks for Covid-19 drug-repurposing} \newblock arXiv:2007.10261 [Preprint]. 2020 [cited 2020 Nov 20]. Available from: 
\url{https://arxiv.org/abs/2007.10261}.

\bibitem{SIGN}
{Rossi E, Frasca F, Chamberlain B, Eynard D, Bronstein M, Monti F.}
\newblock {SIGN: Scalable Inception Graph Neural Networks}
\newblock arXiv:2004.11198 [Preprint]. 2020 [cited 2020 Nov 20]. Available from: 
\url{https://arxiv.org/abs/2004.11198}.

\bibitem{DRKG}
{Ioannidis VN, Song X, Manchanda S, Li M, Pan X, Zheng D, et al.}
\newblock {DRKG - Drug Repurposing Knowledge Graph for Covid-19} 
\newblock {2020} [cited 2020 Nov 20]. Database: Github [Internet]. Available from: \url{https://github.com/gnn4dr/DRKG/}.

\bibitem{Gordon}
{Gordon DE, Jang GM, Bouhaddou M, Xu J, Obernier K, White KM, et al.}
\newblock {A SARS-CoV-2 protein interaction map reveals targets for drug repurposing}.
\newblock {Nature}. {2020};{583}:{459--468}.

\bibitem{Multiorgan}
{Zaim S, Chong JH, Sankaranarayanan V, Harky A. }
\newblock {COVID-19 and multi-organ response}.
\newblock {Curr Prob Cardiology}. {2020};{45}({8}):{100618}.

\bibitem{AI_1}
{Zhang W, Zhao Y, Zhang F, Wang Q, Li T, Liu Z, et al.}
\newblock {The use of anti-inflammatory drugs in the treatment of people with severe coronavirus disease 2019 (COVID-19): The experience of clinical
immunologists from China}.
\newblock Clin Immunol. {2020};{214}:{108393}.

\bibitem{AI_2}
{Channappanavar R, Perlman S. }
\newblock {Pathogenic human coronavirus infections: causes and consequences of cytokine storm and immunopathology}.
\newblock Semin Immunopathol. {2017};{39}({5}):{529--539}.

\bibitem{AI_3}
{Chousterman BG, Swirski FK, Weber GF.}
\newblock {Cytokine storm and sepsis disease pathogenesis}.
\newblock Semin Immunopathol. {2017};{39}({5}):{517--528}.

\bibitem{Nucleotide}
{Jockusch S, Tao C, Li X, Anderson TK, Chien M, Kumar S, et al.}
\newblock {Library of Nucleotide Analogues Terminate RNA Synthesis Catalyzed by Polymerases of Coronaviruses Causing SARS and COVID-19}.
\newblock Antiviral Res. {2020};{180}:{104857}.

\bibitem{Ivermectin}
{Caly L, Druce JD, Catton MG, Jans DA, Wagstaff KM.}
\newblock {The FDA-approved drug ivermectin inhibits the replication of SARS-CoV-2 in vitro}.
\newblock Antiviral Res. {2020};{178}:{104787}.

\bibitem{Statin}
{Fajgenbaum DC, Rader DJ.}
\newblock {Teaching old drugs new tricks: statins for COVID-19?}
\newblock Cell Metab. {2020};{32}({2}):{145--147}.

\bibitem{Statin_1}
{Shin YH, Min JJ, Lee JH, Kim EH, Kim GE, Kim MH, et al.}
\newblock {The effect of fluvastatin on cardiac fibrosis and angiotensin-converting enzyme-2 expression in glucose-controlled diabetic rat hearts}.
\newblock Heart Vessels. {2016};{32}({5}):{618--627}.

\bibitem{ACE_1}
{Zhang H, Penninger JM, Li Y, Zhong N, Slutsky AS.}
\newblock {Angiotensin-converting enzyme 2 (ACE2) as a SARS-CoV-2 receptor: molecular mechanisms and potential therapeutic target}.
\newblock Intensive Care Med. {2020};{46}({4}):{586--590}.

\bibitem{ACE_2}
{Zhang XJ, Qin JJ, Cheng X, Shen L, Zhao YC, Yuan Y, et al.}
\newblock {In-hospital use of statins is associated with a reduced risk of mortality among individuals with COVID-19}.
\newblock Cell Metab. {2020};{32}({2}):{176--187}.

\bibitem{YFV}
{Rega Institute, Belgian University KU Leuven}. 
\newblock {KU Leuven Breakthrough as Modified Yellow Fever Virus Destroys COVID-19 in Preclinical Animal Research: Clinical Trials Next}.
\newblock News article. {2020} [cited 2020 Nov 19]. Available from: \url{https://www.trialsitenews.com/ku-leuven-breakthrough-as-modified-yellow-fever-virus-destroys-covid-19-in-preclinical-animal-research-clinical-trials-next/}.

\bibitem{Mercaptopurine}
{Chen X, Chou CY, Chang GG.}
\newblock {Thiopurine analogue inhibitors of severe acute respiratory syndrome-coronavirus papain-like protease, a deubiquitinating and deISGylating enzyme}.
\newblock Antivir Chem Chemother. {2009};{19}({4}):{151--156}.

\bibitem{Riboflavin}
{Ragan I, Hartson L, Pidcoke H, Bowen R, Goodrich R.}
\newblock {Pathogen reduction of SARS-CoV-2 virus in plasma and whole blood using riboflavin and UV light}.
\newblock {PLOS One}. {2020};{15}({5}):{e0233947}.

\bibitem{TCM}
{Weng JK. }
\newblock {Plant Solutions for the COVID-19 Pandemic and Beyond: Historical Reflections and Future Perspectives}.
\newblock Mol Plant. {2020};{13}:{803--807}.

\bibitem{Drugbank}
{Wishart DS, Feunang YD, Guo AC, Lo EJ, Marcu A, Grant JR, et al.}
\newblock {Drugbank 5.0: a major update to the Drugbank database for 2018}.
\newblock {Nucleic Acids Res.} {2017};{46}({D1}):{D1074--D1082}.

\bibitem{Hetionet}
{Himmelstein DS, Lizee A, Hessler C, Brueggeman L, Chen SL, Hadley D, et al.}
\newblock {Systematic integration of biomedical knowledge prioritizes drugs for repurposing}.
\newblock {eLife}. {2017};{6}:{e26726}.

\bibitem{GNBR}
{Percha B, Altman RB. }
\newblock {A global network of biomedical relationships derived from text}.
\newblock {Bioinformatics}. {2018};{34}({15}):{2614--2624}.

\bibitem{STRING}
{Szklarczyk D, Gable AL, Lyon D, Junge A, Wyder S, Huerta-Cepas J, et al.}
\newblock {STRING v11: protein–protein association networks with increased coverage, supporting functional discovery in genome-wide experimental datasets}.
\newblock {Nucleic Acids Res.} {2019};{47}({D1}):{D607--D613}.

\bibitem{IntAct}
{Orchard S, Ammari M, Aranda B, Breuza L, Briganti L, Broackes-Carter F, et al.}
\newblock {The MIntAct project—IntAct as a common curation platform for 11 molecular interaction databases}.
\newblock {Nucleic Acids Res.} {2014};{42}({D1}):{D358--D363}.

\bibitem{DGIdb}
{Cotto KC, Wagner AH, Feng YY, Kiwala S, Coffman AC, Spies G, et al.}
\newblock {GIdb 3.0: a redesign and expansion of the drug–gene interaction database}.
\newblock {Nucleic Acids Res.} {2018};{46}({D1}):{D1068--D1073}.

\bibitem{Bronstein}
{Bronstein MM, Bruna J, LeCun Y, Szlam A, Vandergheynst P.}
\newblock {Geometric deep learning: going beyond euclidean data}.
\newblock IEEE Signal Process Mag. {2017};{34}({4}):{18--42}.

\bibitem{GCN}
{Kipf TN, Welling M. }
\newblock {Semi-supervised classification with graph convolutional networks}.
\newblock arXiv:1609.02907 [Preprint]. 2016 [cited 2020 Nov 20]. Available from: \url{https://arxiv.org/abs/1609.02907}.

\bibitem{GraphSAGE}
{Hamilton W, Ying Z, Leskovec J.}
\newblock {Inductive representation learning on large graphs}.
\newblock Advances in NeurIPS. 2017 (pp. 1024-1034).

\bibitem{GAT}
{Veli\v{c}kovi\'{c} P, Cucurull G, Casanova A, Romero A, Lio P, Bengio Y.}
\newblock {Graph attention networks}.
\newblock arXiv:1710.10903 [Preprint]. 2017 [cited 2020 Nov 20]. Available from: \url{https://arxiv.org/abs/1710.10903}.

\end{thebibliography}
\end{document}